\documentclass[runningheads]{llncs}

\usepackage{accv}

\usepackage{accvabbrv}

\usepackage{graphicx}
\usepackage{booktabs}
\usepackage{multirow,comment}
\usepackage{booktabs}
\usepackage{sidecap}
\usepackage{array}
\usepackage{stackengine}

\newsavebox\mybox
\newcommand\Includegraphics[2][]{\sbox{\mybox}{%
  \includegraphics[#1]{#2}}\abovebaseline[-.5\ht\mybox]{%
  \addstackgap{\usebox{\mybox}}}}

\usepackage[accsupp]{axessibility}  

\usepackage[pagebackref,breaklinks,colorlinks,citecolor=accvblue]{hyperref}

\usepackage{orcidlink}

\newcommand*{\our}{\texttt{Chameleon}}

\begin{document}

\title{Chameleon: Images Are What You Need For Multimodal Learning Robust To Missing Modalities} 

\titlerunning{Chameleon}

\author{Muhammad Irzam Liaqat\inst{1}\orcidlink{0009-0009-1265-7337} \and
Shah Nawaz\inst{2}\orcidlink{0000-0002-7715-4409} \and
Muhammad Zaigham Zaheer\inst{3}\orcidlink{0000-0001-8272-1351} \and
Muhammad Saad Saeed \inst{4}\orcidlink{0000-0002-0893-9499} \and
Hassan Sajjad \inst{5}\orcidlink{0000-0002-8584-6595} \and
Tom De Schepper \inst{6}\orcidlink{0000-0002-2969-3133} \and
Karthik Nandakumar  \inst{3}\orcidlink{0000-0002-6274-9725} \and
Muhammad Haris Khan \inst{3}\orcidlink{0000-0001-9746-276X} \and
Markus Schedl  \inst{2,7}\orcidlink{0000-0003-1706-3406}
}

\authorrunning{Muhammad Irzam Liaqat et al.}

\institute{IMT School for Advanced Studies of Lucca, \and
Institute of Computational Perception, Johannes Kepler University\\ \and
Mohamed bin Zayed University of Artificial Intelligence \and
Swarm Robotics Lab NCRA, University of Engineering and Technology, Taxila \and 
Dalhousie University \and 
Interuniversity Microelectronics Centre (IMEC), \and 
Human-centered AI Group, AI Lab, Linz Institute of Technology \\
\email{irzam.liaqat@imtlucca.it},
\email{\{shah.nawaz,markus.schedl\}@jku.at}
}

\maketitle

\begin{abstract}
Multimodal learning has demonstrated remarkable performance improvements over unimodal architectures. However, multimodal learning methods often exhibit deteriorated performances if one or more modalities are missing. This may be attributed to the commonly used multi-branch design containing modality-specific streams making the models reliant on the availability of a complete set of modalities. 
In this work, we propose a robust textual-visual multimodal learning method, \our{}, that completely deviates from the conventional multi-branch design. To enable this, we present the unification of input modalities into one format by encoding textual modality into visual representations.
As a result, our approach does not require modality-specific branches to learn modality-independent multimodal representations making it robust to missing modalities.
Extensive experiments are performed on four popular challenging datasets including Hateful Memes, UPMC Food-101, MM-IMDb, and Ferramenta. 
\our{} not only achieves superior performance when all modalities are present at train/test time but also demonstrates notable resilience in the case of missing modalities. 
\keywords{Multimodal learning \and Vision and other modalities \and Missing modalities}
\end{abstract}

\section{Introduction}
\label{sec:intro}
\begin{SCfigure}[][t]
    \includegraphics[width=8.0cm]{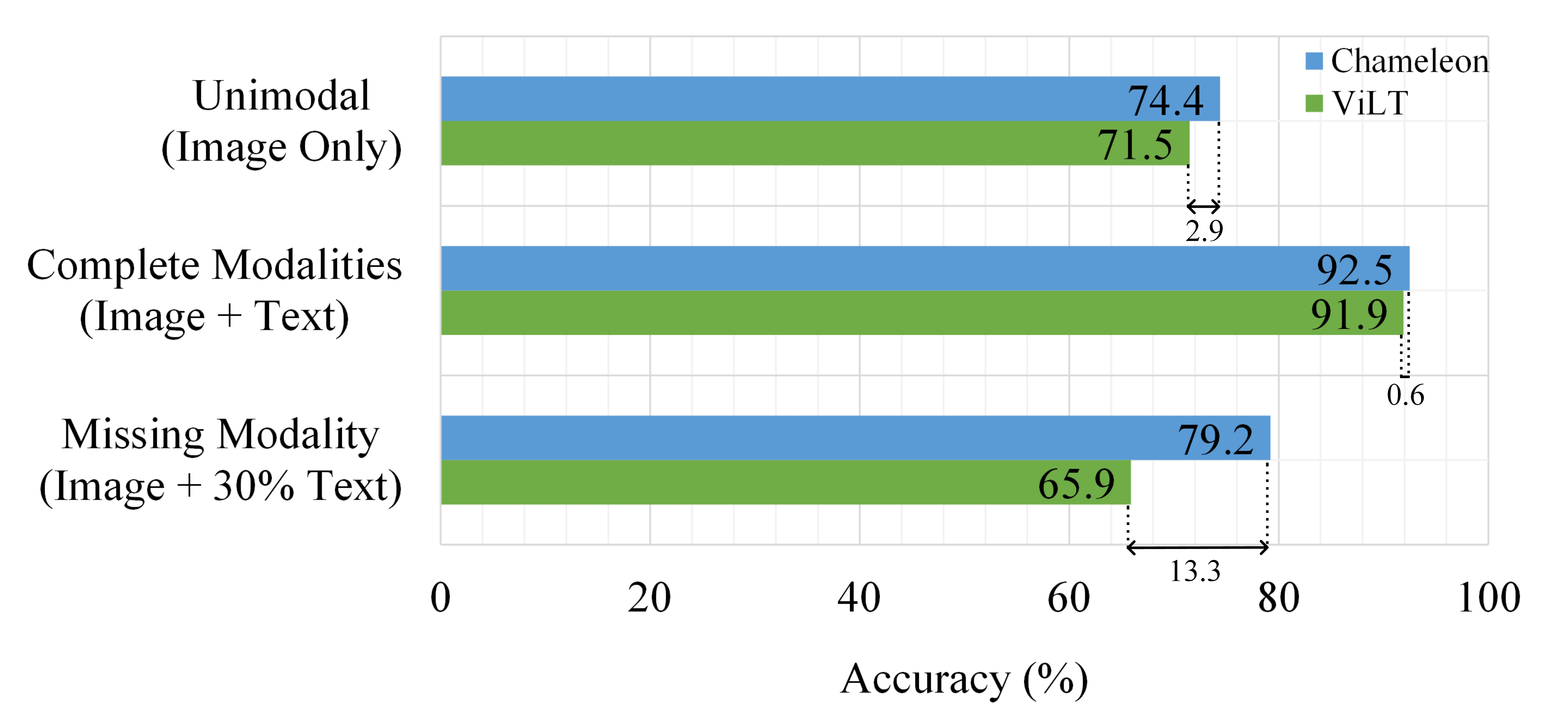}
  \caption{Comparisons of \our{} with ViLT using Food-$101$. \our{} demonstrates better multimodal performance as well as retains superior performance when textual modality is $70$\%  (30\% available) missing at test time. }
  \label{fig:intro}
\end{SCfigure}

Recent years have seen a surge in the use of multimodal data for various applications. 
For instance, users combine text, image, audio, or video modalities to sell a product over an e-commerce platform or express views on social media platforms. 
Two or more of these modalities are often combined in a multi-branch network to solve different tasks such as multimodal classification~\cite{kiela2018efficient,kiela2020hateful}, cross-modal retrieval~\cite{wang2016learning}, cross-modal verification~\cite{nagrani2018seeing}, multimodal named entity recognition~\cite{moon1078,arshad2019aiding}, visual question answering~\cite{anderson2018bottom,fukui2016multimodal}, image captioning~\cite{vinyals2015show}, semantic relatedness~\cite{kiela2014learning}, and multimodal machine translation~\cite{specia2016shared,elliott2016multi30k}.
In all these tasks, multimodal methods can experience scenarios where some modalities are missing, e.g., due to failures in data acquisition pipelines.
Several researchers have recently concluded that multimodal learning is not \textit{inherently} robust to missing modalities and can result in a significantly deteriorated performance when modalities are missing~\cite{ma2022multimodal,ma2021smil,lee2023multimodal}.
For example, as seen in Fig.~\ref{fig:intro}, ViLT \cite{kim2021vilt}, a state-of-the-art (SOTA)textual-visual Transformer, demonstrates a drop in performance of $28.3$\% when only $30$\% of the textual modality is available (i.e., $70$\% missing) at test time. 
The performance deterioration may be attributed to the fundamental design of multi-branch methods \cite{kim2021vilt,lu2019vilbert,wang2016learning,nagrani2018learnable} for multimodal learning that strives to project each modality to a joint embedding space. As such, this makes the model \textit{inherently} reliant on the availability of a complete set of modalities~\cite{ma2022multimodal}.

To address the shortcomings of existing multi-branch networks, we propose \our{}, a multimodal learning method operating on textual-visual modalities that encodes text embeddings as an image.
This way, the network is able to process either of the two inputs, visual or encoded text, as well as both together, all as images. 
Since this encoding is obtained at the input level, the weights are shared across modalities, thereby making the model independent of modality-specific branches.
To perform training, we propose two variants of \our{}, illustrated in Fig.~\ref{fig:main_arch}c and Fig.~\ref{fig:main_arch}d. While more details about these variants are discussed in Section \ref{section:overall_framework}, when a modality is missing during training or testing \our{} shifts its attention to the available modality to output the correct prediction scores.
%
We evaluate \our{} on four textual-visual datasets, including UPMC Food-$101$~\cite{wang2015recipe}, Hateful Memes~\cite{kiela2020hateful}, MM-IMDb~\cite{arevalo2017gated}, and Ferramenta~\cite{gallo2017multimodal}.
It outperforms SOTA multimodal methods on complete modalities and demonstrates superior robustness against missing modalities at train and test times.
For instance, as seen in Fig.~\ref{fig:intro}, compared to ViLT~\cite{kim2021vilt}, \our{} consistently demonstrates better classification accuracy in both missing modality and complete modality scenarios. 
Notably, \our{} outperforms ViLT in the missing modality setting by a significant margin of $13.3$\%.

\noindent The key contributions of our work are as follows:
\begin{enumerate}
\item We propose \our, a textual-visual multimodal method \textit{inherently} robust to missing modalities.
\item We explore an encoding approach that transforms textual information into visual representations to carry out robust multimodal training. 
\item We explore the general applicability of our approach towards robustness to missing modalities on Convolutional Neural Networks (CNNs) as well as Vision Transformers (ViT). To the best of our knowledge, \our{} is the first work that explores the robustness to missing modality across multiple learning paradigms including CNNs and ViTs.
\item  A wide range of experiments performed on various textual-visual datasets 
under different unimodal, multimodal, and missing modality settings during training and testing demonstrate the significance of \our{}, making it a defacto approach for missing modality related applications.
\end{enumerate}

\section{Related Work}
\label{sec:related-work}

\subsection{Multimodal Learning and Missing Modality}
Multiple modalities including texts and images often contain complementary information about a common subject. The goal of multimodal learning is to leverage complementary information across modalities to improve the performance of various machine learning tasks.
Each task may be different from the other, however, the underlying objective remains the same: to learn joint representations across multiple modalities~\cite{baltruvsaitis2018multimodal,xu2023multimodal}. 
Existing multimodal methods employ multi-branch networks to learn joint representations by minimizing the distance between the representations of multiple modalities~\cite{nagrani2018learnable,saeed2022fusion,kim2018learning,arevalo2017gated,vielzeuf2018centralnet,kiela2018efficient,kim2021vilt,lu2019vilbert,radford2021learning,nagrani2018seeing}.
Such methods have achieved remarkable performance using modality-complete data~\cite{he2017fine,arevalo2017gated, vielzeuf2018centralnet,kiela2018efficient,yang2019exploring,kiela2020hateful,kim2021vilt}. 

However, multimodal methods suffer from performance deterioration if some modalities are absent either during training or testing~\cite{ma2022multimodal,lee2023multimodal,zhang2022multimodal,suo2019metric}. 
Considering the importance of multimodal methods, recent years have seen a surge in tackling the missing modality problem~\cite{ma2021smil,ma2022multimodal,lee2023multimodal,zhang2022m3care,wang2022m2r2}. For example, Ma et al.~\cite{ma2022multimodal} improved the robustness of Transformer models via multi-task optimization.
More recently, Lee et al.~\cite{lee2023multimodal} introduced missing-modality-aware prompts which can be plugged into multimodal Transformers to handle missing modality.  
\our{} is related to these as we also address the missing modality problem. However, we approach the problem by converting textual modality into a visual format to train a visual network. This makes our approach independent of any specific component such as missing-modality-aware prompts towards tackling the missing modality problem.

\subsection{Rendering Non-visual Modalities into Images}
Recently, some researchers have explored the idea of transforming text information into a visual format to perform visual recognition task. For example,
Gallo et al.~\cite{gallo2017semantic} leverage Word$2$Vec word embeddings to reconstruct the semantics associated with text as an image to train a visual network for text classification.
Salesky et al.~\cite{salesky2021robust} render the raw text directly into visual format, divide the rendered image into overlapping slices, and produce representations with optical character recognition to train a machine translation model.
Similarly, Rust et al.~\cite{rust2022language} propose a Masked Autoencoding Visual Transformer to reconstruct the pixels in masked image patches to train a language model.
More recently, Tschannen et al.~\cite{tschannen2023clippo} proposed CLIPPO that renders text information as images to train a pure pixel-based model performing multimodal learning. 
All approaches rendering text into visual format are, in essence, related to \our{} as we also encode text into visual format. 
However, instead of the common approach of transforming raw text directly into images, we use word embeddings to encode textual information as color-coded pixels which is found to be effective in our experiments for the missing modality problem.
Further, we explore the utilization of joint visual representations of multimodal inputs to train a robust multimodal classification method as well as study its resilience against missing modalities. 
In addition, although the scope of our approach is textual-visual data, we acknowledge the existing approaches that encode audio modality into visual format to train on a given task~\cite{nagrani2017voxceleb,xie2019utterance}.

\subsection{Mixing Multiple Input Instances}
The second variant of our training strategy, \textit{fused}, is loosely inspired by the recent progress in image augmentation \cite{zhang2017mixup,yun2019cutmix} where two input images are mixed in such a way that the network is trained to predict both class labels at the same time. Such approaches have been proven effective towards better generalization, i.e.,~when only one image is provided to the network at test time, the model is able to classify it correctly. We extend this idea to improve the multimodal performance as well as robustness of \our{} towards missing modality by \textit{augmenting} the encoded text with its corresponding image to create a fused multimodal image. The network is then trained using this fused image to predict the class labels.  

\begin{figure*}
\centering
\includegraphics[width=0.99\linewidth]{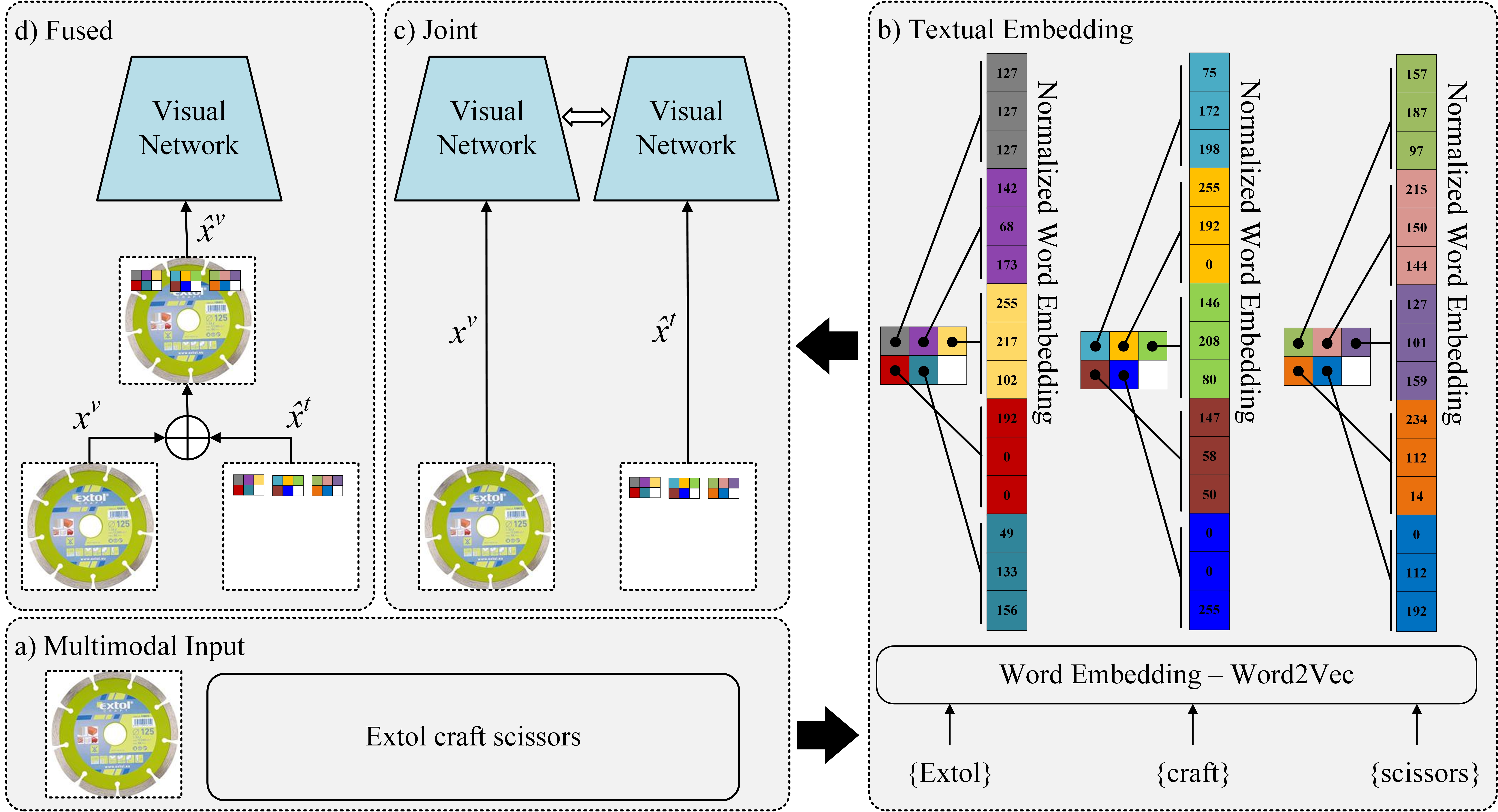}
\caption{The overall architecture of \our{}. (a) Multimodal input consists of images and text. (b) Word embeddings are encoded as an image. Example of encoding three words: `Extol', `craft', and `scissors' is provided with embedding length $15$. (c) The \textit{joint} variant of \our{} in which visual and encoded text inputs are fed to a weight-sharing visual network in a sequential fashion. 
(d) The \textit{fused} variant of \our{} in which a {fused} image created by fusing both visual and encoded text inputs are fed to the network.
}
\label{fig:main_arch}
\end{figure*}

\section{Methodology}
\label{section:overall_framework}
We present \our{}, a multimodal learning method consisting of textual-visual modalities that encodes text embeddings as images and subsequently trains an image classification network to learn joint representation. 
Our approach is built on the intuition that learning on a common input format with weight sharing across multiple modalities results in a robust multimodal method resilient to missing modalities.  
In order to carry out the training, we explore two different input approaches namely \textit{joint} and \textit{fused}. 
Fig.~\ref{fig:main_arch} summarizes \our{} whereas 
each of the components is discussed subsequently in this section.

\subsection{Problem Formulation}
Formally, given $\mathcal{D}=\{(x_{i}^{t},x_{i}^{v})\}_{i=1}^{N}$ is the training set where $N$ is the number of pairs of modality $t$ (text) and modality $v$ (visual) and $x_{i}^{t}$ and $x_{i}^{v}$ are individual modality samples of the $i^{th}$ instance respectively. 
Moreover, each pair $(x_{i}^{t},x_{i}^{v})$ has a class label $y_{i}$.
Typical existing multimodal methods take multiple modalities as input by using a multi-branch network $\mathcal{C}_m$ to perform the classification task:

\begin{equation}
    \tilde{y}_i = \mathcal{C}_m (\{x^t, x^v\}, y_i)
    \label{eq:existingclassifiers}
\end{equation}

\noindent Such a multi-branch configuration requires modality-complete data to perform a given task and, as seen in Fig. \ref{fig:intro}, missing modality may result in a significant performance deterioration~\cite{kim2021vilt,ma2022multimodal}.

In \our{}, we propose to encode text input as image and subsequently perform the task entirely in the visual domain, both during 
training and testing.
In other words, a visual network is trained on multimodal data consisting of images and encoded texts through a single interface of visual information.
We hypothesize that unifying the modalities to an identical input format will 
make the model independent of modality-specific branches and will result in a method robust to missing modalities.
Therefore, in our approach,  Eq.~\eqref{eq:existingclassifiers} takes the following form:

\begin{equation}
     \tilde{y}_i = \mathcal{C}_v (\{\hat{x}^t, x^v\}, y_i),
\end{equation}
\noindent where $\mathcal{C}_v$ is a visual classifier, $\hat{x}^t = \mathcal{E} (x^t)$, and
$\mathcal{E}$ is the encoding scheme that transforms $x^t$ into its visual format $\hat{x}^t$. The encoding scheme is discussed next.

\subsection{Text Encoding Scheme}
 \label{subsec:encoding}
Mikolov et al. \cite{Mikolov:2013:ICLR} showed that words group together in high dimensional space based on various linguistic knowledge such as semantics, morphology, and syntax. Moreover, Dalvi et al. \cite{dalvi2022discovering} showed the existence of multifaceted groups in the latent space, i.e., groups that are formed based on more than one relationship among words. We hypothesize that the knowledge of these relationships is represented in 
the feature vectors.
Explicitly considering all 
features
in the form of an image will enable the preservation of the rich knowledge learned in the network. Building upon this, we use 
combinations of three consecutive values in the feature vector as RGB pixel blocks. 
Formally,
%
for a given pair of $(x_{i}^{t},x_{i}^{v})$, the encoding scheme takes a word in $x_{i}^{t}$ and converts it into a feature vector.
The feature vector is then normalized to assume values in the interval $[0\dots255]$.
Afterward, consecutive three elements
in the normalized feature vector are considered as one RGB value.
The process is repeated for each word in $x_{i}^{t}$ to obtain the final visual encoding $\hat{x}_{i}^{t}$.
Fig.~\ref{fig:main_arch}b shows examples of encoding input words into an RGB sequence. 
To train the network, we explore two possible configurations:\\
\noindent \textbf{Joint.}
The encoded text information ($\hat{x}_{i}^{t}$) is patched onto a blank image to create a joint visual representation (Fig.~\ref{fig:main_arch}c). This results in an additional image containing only the encoded text as visual information.
For training, the two images are input in a sequential fashion to a weight-shared visual network.\\
\noindent \textbf{Fused.}
The encoded text ($\hat{x}_{i}^{t}$) is patched onto its corresponding input image ($x_{i}^{v}$) to create a fused visual representation (Fig.~\ref{fig:main_arch}d)
containing both modalities. The training of the visual classifier ($\mathcal{C}_v$) is carried out in a typical fashion. 

\begin{figure}[t]
\footnotesize
\centering 
\begin{tabular}{  p{0.7cm} >{\centering\arraybackslash}m{2cm} >{\centering\arraybackslash}m{2cm}  >{\centering\arraybackslash}m{2cm} >{\centering\arraybackslash}m{2cm} >{\centering\arraybackslash}m{2cm} >{\centering\arraybackslash}m{2cm} }      
\hline                                      
& & Baseline & \multicolumn{3}{c}{\our} \\ [0.5ex] 
\hline                                      
 & a. Orig. Image & b.    Unimodal & c. Multimodal &  d. Missing Text &  e. Missing Image\\ [0.5ex] 
\hline
\multirow{2}{*}{\rotatebox[origin=c]{90}{Ice Cream}}
& \Includegraphics[height=0.73in]{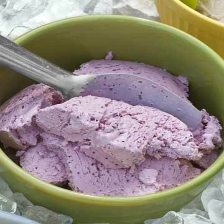}
& \Includegraphics[height=0.73in]{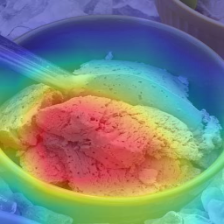}
& \Includegraphics[height=0.73in]{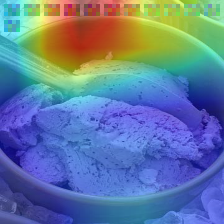} 
& \Includegraphics[height=0.73in]{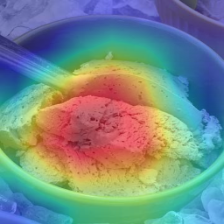} 
& \Includegraphics[height=0.73in]{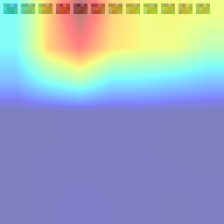} 
\\ [1ex] 
& & 0.60 & 0.99 & 0.99 &  1.0\\  
\hline
\multirow{2}{*}{\rotatebox[origin=c]{90}{Caser Salad}}
& \Includegraphics[height=0.73in]{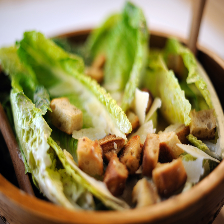}
& \Includegraphics[height=0.73in]{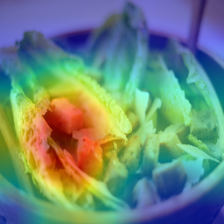}
& \Includegraphics[height=0.73in]{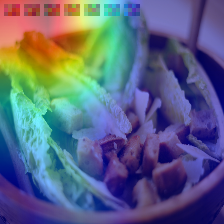} 
& \Includegraphics[height=0.73in]{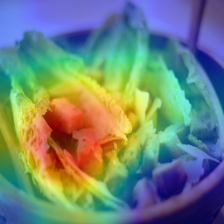} 
& \Includegraphics[height=0.73in]{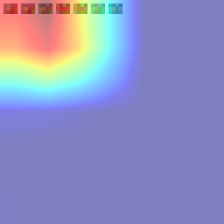} 
\\ [1ex] 
& & 0.98 & 0.99 & 0.99 &  1.0\\  

\hline

\multirow{2}{*}{\rotatebox[origin=c]{90}{Baklava}}
& \Includegraphics[height=0.73in]{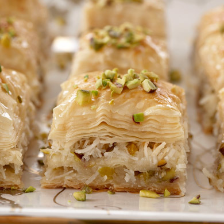}
& \Includegraphics[height=0.73in]{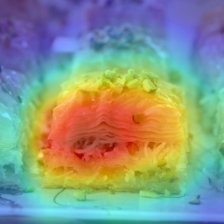}
& \Includegraphics[height=0.73in]{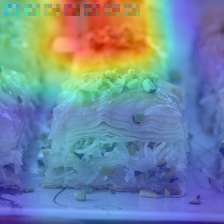} 
& \Includegraphics[height=0.73in]{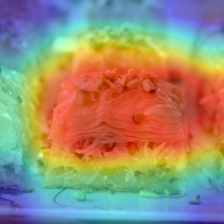} 
& \Includegraphics[height=0.73in]{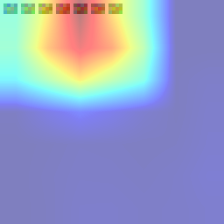} 
\\ [1ex] 
& & 0.98 & 0.99 & 0.99 &  0.99\\
\hline
\end{tabular}
\caption{
Grad-CAM visualizations of selected images from UPMC Food-$101$: (a) Orignal image randomly selected from the test set. (b) Unimodal image only training and testing. (c) Multimodal training and testing. (d) Multimodal training; testing on image only, i.e., 100\% text missing. (e) Multimodal training; testing on text only, i.e., 100\% image missing. As seen, in unimodal image-only training (b), the model focuses on distinct features of the object. With our multimodal training (c), the model not only retains its focus on the object but also includes encodings representing text modality to make the predictions. When the text modality (d) or image modality (e) is missing during testing, the model focuses on the available modality to make accurate final predictions demonstrating the success of \our{} in training the multimodal method robust to missing modalities.}
\label{fig:new_gradcam}
\end{figure}

\subsection{Robustness to Missing Modalities}
\our{} inherently learns shared representations across modalities, bringing them into a similar latent space. This results in a richer representation that is complemented by both modalities and yields a robust model when one of the modalities is missing. In other words, when a modality is missing, the available modality enables the use of the multimodal knowledge from the shared latent space to make a prediction. 
To verify this, we visually compare the activation maps~\cite{selvaraju2017grad} of various different training/testing configurations of \our{} in Fig.~\ref{fig:new_gradcam}, including image-only unimodal training/testing, multimodal training/testing, multimodal training with missing text modality testing, and multimodal training with missing image modality testing.
In image-only unimodal training and testing (Fig.~\ref{fig:new_gradcam}b), as expected, the model focuses on the distinct features of the object to predict the classification score. 
For the case of complete modality during training and testing (Fig.~\ref{fig:new_gradcam}c), \our{} not only retains its focus on parts of the object but also attends to the portions of the image where text modality is encoded. 

In the case of missing modality at test time, a multimodal architecture robust to missing modalities should ideally be able to retain the focus on the available modality and behave like an unimodal network. We observe this case in Fig.~\ref{fig:new_gradcam}d for \our{}, when the text modality is missing at test time. 
As seen, the model shifts its focus on the object itself and behaves similarly to the unimodal network.
Likewise, when the image modality is missing at test time (Fig.~\ref{fig:new_gradcam}d), \our{} shifts its focus to the encoded textual representation.
These visualizations highlight the internal working of \our{} in successfully learning multimodal representations while demonstrating resilience against missing modalities.

\section {Experiments and Analysis}
\label{section:experi}
We evaluate \our{} on the multimodal classification task using four popular and challenging datasets including UPMC Food-$101$~\cite{wang2015recipe}, Hateful Memes~\cite{kiela2021hateful}, MM-IMDb~\cite{vielzeuf2018centralnet}, and Ferramenta~\cite{gallo2017multimodal}.
Recently, Ma et al.\cite{ma2022multimodal} and Lee et al.~\cite{lee2023multimodal} introduced comprehensive protocols to study the missing modality problem on textual-visual data at train and test time. 
Specifically, multimodal methods are trained/tested on different predefined levels of missing modality to observe deterioration.
We conduct experiments using these settings to make results comparable.
Moreover, an extensive ablation study is performed to evaluate different design choices of \our{}.

\subsection{Datasets}
In addition to the three textual-visual datasets used in the existing literature on missing modality \cite{ma2022multimodal,lee2023multimodal}, we extend our study by selecting another widely popular and challenging multimodal dataset, Ferramenta~\cite{gallo2017multimodal}, which is curated to resolve ambiguities among visual samples by using the textual modality. 

\noindent \textbf{UPMC Food-101.}  It is a multimodal dataset consisting of textual and visual modalities. 
The dataset is crawled from the web and each entry consists of an image and the HTML web page on which it was found.
It contains $90,704$ pairs divided into $101$ food classes, and comes with a predefined $75$/$25$ train/test split. \\
\noindent \textbf{Hateful Memes.} It is a multimodal dataset containing meme images and their respective textual contents as the two modalities with binary labels and is developed with the aim of identifying hate speech in memes. 
The dataset contains $10,000$ memes.\\
\noindent \textbf{MM-IMDb.} It is a multimodal dataset containing textual-visual modalities to predict the genres of a movie. 
It contains $25,956$ image-text pair and $23$ classes.\\
\noindent \textbf{Ferramenta.} It consists of $88,010$ adverts belonging to $52$ different product classes extracted from an e-commerce platform representing product images and their textual specifications. The data is divided into $66,141$ instances for train and $2,186$ instances for test. 

\subsection{Evaluation Metrics}
Following existing SOTA methods on complete and missing modalities~\cite{lee2023multimodal,ma2022multimodal,vielzeuf2018centralnet,wang2015recipe,kiela2020hateful}, we report classification accuracy for UPMC Food-$101$ and Ferramenta, area under the receiver operating characteristic (AUROC) for Hateful Memes and F$1$ Macro for MM-IMDb.

\subsection{Implementation Details}
We consider both CNN (ResNet-$101$~\cite{he2016deep}) and ViT~\cite{dosovitskiy2020image} ((vit-base-patch$16$-$224$) as visual networks. 
Both variants including joint and fused are also evaluated individually by carrying out separate training on each network. 
Based on the detailed empirical studies reported in Sections~\ref{subsec:analysis}, unless stated otherwise, \textit{ViT with joint representation} is our default choice of architecture for training and testing.  
We train Word$2$Vec for every dataset and the embedding length is set to $36$ for encoding words to visual format.
Rather than training from scratch, we fine-tune both ResNet-$101$ and ViT pre-trained on ImageNet.

\subsection{Evaluations Under Complete Modalities Setting} 

To evaluate whether \our{} is capable of learning from multiple input modalities, we first evaluate when complete modalities are present during training and testing. 
As seen in Table~\ref{tab:complete}, \our{} achieves SOTA performance on three of the four datasets including UPMC Food-$101$, Hateful memes, and Ferramenta. 
More specifically, \our{} achieves classification performances of $92.5$\%, $73.9$\% and $96.6$\%, respectively, outperforming other state-of-the-art methods. 
Discussion about the performance of \our{} on MM-IMDb dataset is provided in the next section.
Overall, the results demonstrate that \our{} is capable of learning from multiple modalities on par with the existing SOTA multimodal methods. 

\begin{table}[t]
\footnotesize

\caption{Comparison of \our{} with SOTA multimodal methods on UPMC Food-$101$, Hateful Memes, MM-IMDb, and Ferramenta datasets using modality-complete data.
Best results are bold; second best are underlined.}
\resizebox{0.23\linewidth}{!}{
\begin{tabular}[t]{p{2.8cm}|c}
\hline
\multicolumn{2}{c}{\textbf{UPMC Food-$101$}~\cite{wang2015recipe}} \\
\hline\hline
Method & Acc. \\
\hline
Wang et al.~\cite{wang2015recipe}    & 85.1 \\
\hline
Fused Rep.~\cite{nawaz2018learning}    & 85.7 \\
\hline
CLIPPO~\cite{tschannen2023clippo}      & 91.2 \\
\hline
CentralNet~\cite{vielzeuf2018centralnet} & 91.5 \\
\hline
 ViLT~\cite{kim2021vilt}      & 91.9 \\
\hline
Ma et al.~\cite{ma2022multimodal}    & 92.0 \\
\hline
 MMBT~\cite{kiela2019supervised}    & \underline{92.1} \\
\hline
BL~\cite{gallo2020image}   & \textbf{92.5} \\
\hline

\our{}  & \textbf{92.5} \\
\hline
\end{tabular}
}
\hfill
\resizebox{0.23\linewidth}{!}{
\begin{tabular}[t]{p{2.8cm}|c}
\hline
\multicolumn{2}{c}{\textbf{Hateful Memes~}~\cite{kiela2020hateful}} \\
\hline\hline
Method  & AUROC \\
\hline
MMBT-G~\cite{kiela2019supervised}     & 67.3 \\
\hline
ViLT~\cite{kim2021vilt}               & 70.2 \\
\hline
Clippo~\cite{tschannen2023clippo}     & 70.7 \\
\hline
Ma et al.\cite{ma2022multimodal}      & 71.8 \\
\hline
MMBT-R~\cite{kiela2019supervised}     & 72.2 \\
\hline
Visual BERT~\cite{li2019simple}       & 73.2 \\
\hline
ViLBERT~\cite{lu2019vilbert}          & \underline{73.4} \\
\hline
\our{}                                & \textbf{73.9} \\
\hline
\end{tabular} 
}
\hfill
\resizebox{0.23\linewidth}{!}{
\begin{tabular}[t]{p{2.3cm}|c}
\hline
\multicolumn{2}{c}{\textbf{MM-IMDb}~\cite{vielzeuf2018centralnet}} \\
\hline
\hline
Method  & F1 Marco \\
\hline 
CBGP~\cite{kiela2018efficient} &  52.9 \\ \hline
CBP~\cite{fukui-etal-2016-multimodal}              & 53.2 \\ \hline
GMU~\cite{arevalo2017gated} &  53.9 \\ \hline
Lee et al.~\cite{lee2023multimodal} & 54.0 \\ \hline
ViLT~\cite{kim2021vilt}& 55.3\\ \hline 
MFAS~\cite{perez2019mfas}&\underline{55.7}\\ \hline
CentralNet~\cite{vielzeuf2018centralnet}&\textbf{56.1} \\ \hline
\our{}  & 51.2 \\
\hline
\end{tabular} }
\hfill
\resizebox{0.23\linewidth}{!}{
\begin{tabular}[t]{p{2.3cm}c}
\hline
\multicolumn{2}{c}{\textbf{Ferramenta}~\cite{gallo2017multimodal}} \\
\hline
\hline
Method  & Acc.\\
\hline 
Ferramenta~\cite{gallo2017multimodal}           &  92.9 \\ \hline
Fused Rep.~\cite{nawaz2018learning}             &  94.8 \\ \hline
IeTF~\cite{gallo2018image}                      &  95.2 \\ \hline
Two-Branch~\cite{saeed2022fusion}               &  96.2 \\ \hline
MHFNet~\cite{yue2023multi}                      & \underline{96.5}\\ \hline
\our{}                                          & \textbf{96.6} \\ \hline
\end{tabular} }

\label{tab:complete}

\end{table}

\begin{table*}[t]
\caption{Comparison of \our{} with different levels of available modality at test time using UPMC-Food-$101$, Hateful Memes, MM-IMDb, and Ferramenta datasets. 
Comparison is provided with ViLT~\cite{kim2021vilt}$^\ast$, Ma et al.~\cite{ma2022multimodal}, CLIPPO\cite{tschannen2023clippo}, \& LLaVA~\cite{liu2023llava}.
$^\ast$ViLT values are taken from Ma et al.~\cite{ma2022multimodal}. 
Boldface and underline denote, respectively, the best and second-best results.
$\dagger$ indicates our implementation.}
\centering
\footnotesize
\begin{tabular}{c|cc|cc|c|c|c|c|c}
\hline
\multirow{2}{*}{Data} & \multicolumn{2}{c|}{Training}  & \multicolumn{2}{c|}{Testing}    & \multirow{2}{*}{ViLT$^\ast$} & \multirow{2}{*}{Ma et al.} & \multirow{2}{*}{CLIPPO$\dagger$}  & \multirow{2}{*}{LLaVA$\dagger$} & \multirow{2}{*}{\our{}}  \\

\cline{2-5}
&  Image & Text & Image & Text &  & & & & \\
 \hline\hline
\multirow{6}{*}{\rotatebox[origin=c]{90}{ Food-101}}
& 100\% & 100\% & 100\% & 100\%     & 91.9      & 92.0        & 91.2 & \underline{92.2}   &  \textbf{92.5} \\
& 100\% & 100\% & 100\% & 90\%      & 88.2      & \underline{90.5}        & 89.6 & 90.3   & \textbf{90.8}   \\
& 100\% & 100\% & 100\% & 70\%      & 80.7      & 87.1        & 86.3 & \underline{87.2}   & \textbf{87.3}    \\
& 100\% & 100\% & 100\% & 50\%      & 73.3      & 82.6    & 81.9 & \underline{83.0}      & \textbf{83.1}     \\
& 100\% & 100\% & 100\% & 30\%      & 65.9      & 77.5   & 78.1 & \underline{78.9}   & \textbf{79.2}      \\
& 100\%  & 100\% & 100\% & 10\%     & 58.4      & 73.3 & 74.1 & \underline{75.4}          & \textbf{76.3}       \\ 
\hline
\multirow{6}{*}{\rotatebox[origin=c]{90}{Hateful Memes}}                       
  & 100\% & 100\% & 100\% & 100\%      & 70.2   & \underline{71.8} & 70.7             &  70.7 &\textbf{73.9} \\
  & 100\% & 100\% & 100\% & 90\%       & 68.8   & 69.7             & 70.3 &  \underline{70.5}  & \textbf{73.6}   \\
  & 100\% & 100\% & 100\% & 70\%       & 65.9   & 66.6             & 70.0 &  \underline{70.3} & \textbf{73.4} \\
  & 100\% & 100\% & 100\% & 50\%       & 63.6   & 63.9             & 69.3 & \underline{69.9} & \textbf{73.1}   \\
  & 100\% & 100\% & 100\% & 30\%       & 60.2   & 61.2             & 69.0 & \underline{69.8} &  \textbf{72.8}  \\
  & 100\%  & 100\% & 100\% & 10\%      & 58.0   & 59.6             & 68.1 & \underline{69.5} & \textbf{72.0}   \\ 
\hline
\multirow{6}{*}{\rotatebox[origin=c]{90}{MM-IMDb}}                       
& 100\% & 100\% & 100\% & 100\%     & \underline{55.3}         & 55.0  & 28.4 & \textbf{59.3} & 51.2 \\
& 100\% & 100\% & 100\% & 90\%      & 51.8      & \underline{53.8}     & 27.5 & \textbf{56.7} & 45.7   \\
& 100\% & 100\% & 100\% & 70\%      & 45.1      & \textbf{52.0}     & 25.6 & \underline{51.3} &43.1    \\
& 100\% & 100\% & 100\% & 50\%      & 38.9      & \textbf{46.6}     & 23.6 & \underline{44.0} & 40.4  \\
& 100\% & 100\% & 100\% & 30\%      & 31.2                  & \textbf{41.8}    & 21.2 & 36.7 & \underline{37.9}   \\
& 100\%  & 100\% & 100\% & 10\%     & 23.1 & \textbf{37.3}     & 18.9          & 27.4  & \underline{35.2}    \\ 

\hline
\multirow{6}{*}{\rotatebox[origin=c]{90}{Ferramenta}}                       
  & 100\% & 100\% & 100\%  & 100\%       & \underline{95.9} & -  & 93.4              & 94.8 & \textbf{96.6} \\
  & 100\% & 100\% & 100\%  & 90\%        & 68.1             & -  & 91.6  & \underline{94.0} & \textbf{95.8}  \\
  & 100\% & 100\% & 100\%  & 70\%        & 60.8             & -  & 89.9  & \underline{93.4} & \textbf{95.4}   \\
  & 100\% & 100\% & 100\%  & 50\%        & 60.4             & -  & 88.7  & \underline{93.0} & \textbf{94.9}    \\
  & 100\% & 100\% & 100\%  & 30\%        & 54.2             & -  & 86.1  & \underline{92.7} & \textbf{94.8}    \\
  & 100\%  & 100\% & 100\% & 10\%        & 51.3             & -  & 84.6  & \underline{92.0} & \textbf{94.6}    \\ 
\hline

\hline
\end{tabular}
\label{tab:percentage_missing}
\vspace{-2em}
\end{table*}


\subsection{Evaluations Under Missing Modalities Setting}
Several researchers have shown that multimodal methods are sensitive to missing modalities~\cite{ma2022multimodal,ma2021smil,lee2023multimodal}. \our{} addresses this issue specifically by converting textual information into visual format. In this section, we provide a detailed comparison with existing SOTA methods by extensively evaluating \our{} on missing modality scenarios, both during training and testing.

\subsubsection{Missing Modalities During Testing.}
A common issue with multimodal methods during deployment is encountering missing modalities at test time due to failures in the data acquisition pipeline. 
Table~\ref{tab:percentage_missing} compares \our{} with other multimodal learning methods:  ViLT~\cite{kim2021vilt}, Ma et al.~\cite{ma2022multimodal}, CLIPPO~\cite{tschannen2023clippo}, and LLaVA~\cite{liu2023llava} for varying amounts of missing modality on UPMC Food-$101$, Hateful Memes, MM-IMDb, and Ferramenta datasets during testing. 
As seen, \our{} outperforms compared methods on UPMC Food-$101$, Hateful Memes, and Ferramenta demonstrating strong resilience against missing modalities. 
For example, in the case when only $10$\% of text modality is available on the UPMC Food-$101$ dataset during testing, \our{} demonstrates an accuracy of $76.3$\%. In comparison, ViLT \cite{kim2021vilt}, Ma et al. \cite{ma2022multimodal}, CLIPPO~\cite{tschannen2023clippo} and LLaVA demonstrate performances of $58.4$\%, $73.3$\%, $74.1$\%, and $75.4$\%, respectively. Similar trends are noticeable on Hateful Memes and Ferramenta datasets.
In the case of MM-IMDb dataset, though the performance of \our{} is lower when all modalities are available, it outperforms ViLT, CLIPPO and LLaVa when a modality is severely missing. Overall, considering all 4 datasets presenting 24 scenarios of complete and missing modality, \our{} outperforms all compared methods in 18 while second best in 2 scenarios. 
Another interesting observation is seen when results of \our{} are compared to CLIPPO
which can be considered a baseline to our method that renders text directly as an image.
Furthermore, we compare results with a large-scale vision and language model (LLaVA), which converts image features to textual tokens. Though the methodology is different compared to \our{}, the essence of converting one modality into the other is similar. Comparing results directly with CLIPPO and LLaVA demonstrates the effectiveness of \our{}.
As seen in Table \ref{tab:percentage_missing},
\our{} performs significantly better in most scenarios which not only signifies the importance of our proposed text encoding scheme for an improved generalization of networks robust to missing modalities but also demonstrates the importance of converting textual modality into visual format.

The overall persistent robustness to missing modality of \our{} compared to existing  methods across various challenging datasets and scenarios is owed to the text encoding scheme used that enables the training on visual inputs. This way, if one modality is missing during testing, \our{} shifts its focus to the available modality, thus yielding higher classification performance. More insights on this are provided in Fig.~\ref{fig:new_gradcam} where Grad-CAM visualizations clearly show the shifting of focus from multimodal data to the available modality for predictions. 
More visual examples are provided in the supplementary material. 

\begin{table*}[t]
\caption{Comparison of \our{} with  ViLT~\cite{kim2021vilt}$^\ast$, Lee et al.~\cite{lee2023multimodal}, and CLIPPO~\cite{tschannen2023clippo} on UPMC Food-$101$~\cite{wang2015recipe}, Hateful Memes~\cite{kiela2020hateful}, and MM-IMDb~\cite{arevalo2017gated} under different missing modality settings during training and testing. 
$^\ast$ViLT values are taken from Lee et al.~\cite{lee2023multimodal}.
Boldface and underline denote, respectively, the best and second best results.
$\dagger$ indicates our implementation.
}
\centering
\footnotesize
\resizebox{0.99\linewidth}{!}{
\begin{tabular}{c|cc|cc|c|c|c|c}
\hline
\multirow{2}{*}{Data}  & \multicolumn{2}{c|}{Training}  & \multicolumn{2}{c|}{Testing}  & \multirow{2}{*}{ViLT~\cite{kim2021vilt}}  & \multirow{2}{*}{Lee et al.~\cite{lee2023multimodal}}  & \multirow{2}{*}{CLIPPO~\cite{tschannen2023clippo}$\dagger$} & \multirow{2}{*}{\our{}}\\
\cline{2-5}
 &Image & Text & Image & Text &  & & &\\
\hline\hline
\multirow{3}{*}{UPMC Food-$101$} 
& 100\%  & 100\%  & 100\% & 100\%         & 91.9   & \underline{92.0}  & 91.2  & \textbf{92.5} \\ 
 & 100\%  & 30\%  & 100\% & 30\%          & 66.3   & \underline{74.5}  & 77.5  & \textbf{79.8}      \\
 & 30\%   & 100\%  & 30\% & 100\%         & 76.7   & \textbf{86.2}     & 83.6  & \underline{83.9}\\
 
 \hline
\multirow{3}{*}{Hateful Memes}   
&100\%  & 100\%  & 100\% & 100\%         & 70.2  & \underline{71.0}  & 70.7              & \textbf{73.9}\\   
& 100\%  & 30\%  & 100\% & 30\%          & 60.8  & 59.1              & \underline{60.9}  & \textbf{72.7} \\ 
& 30\%   & 100\%  & 30\% & 100\%         & 61.6  & \textbf{63.1}     & 58.4              & \underline{62.0}\\ 
\hline               
\multirow{3}{*}{MM-IMDb }        
& 100\%  & 100\%  & 100\% & 100\%   & 46.0  & \textbf{54.0}          & 28.4     & \underline{51.2} \\
& 100\%  & 30\%  & 100\% & 30\%      & 35.1  & \textbf{39.2} & 23.6 & \underline{36.8}\\ 
& 30\%   & 100\% & 30\% & 100\%      & 37.7  & \textbf{46.3}             & 23.3 & \underline{45.1}\\ 
\hline
\end{tabular}
}
\label{tab:traing_missing}
\end{table*}
\subsubsection{Missing Modalities During Training.}
Derived from the motivation that modality can be missing in any data sample, Lee et al.~\cite{lee2023multimodal} recently extended the evaluation protocol by introducing scenarios of missing modality during training and testing, i.e., $30$\% of one modality is available against $100$\% of the other modality at train and test time. 
Table~\ref{tab:traing_missing} compares our approach with existing methods  ViLT~\cite{kim2021vilt}, Lee et al.~\cite{lee2023multimodal}, and CLIPPO~\cite{tschannen2023clippo} using this protocol on UPMC Food-$101$, Hateful Memes, and MM-IMDb datasets. 
In most scenarios, \our{} demonstrates comparable or better performance than existing SOTA methods.
Notably, compared to CLIPPO, \our{} demonstrates robustness to missing modalities during training and testing thus reiterating the importance of our text encoding scheme to train multimodal systems.


\begin{figure}
     \centering
     \begin{subfigure}[b]{0.45\textwidth}
         \centering
         \includegraphics[width=\textwidth]{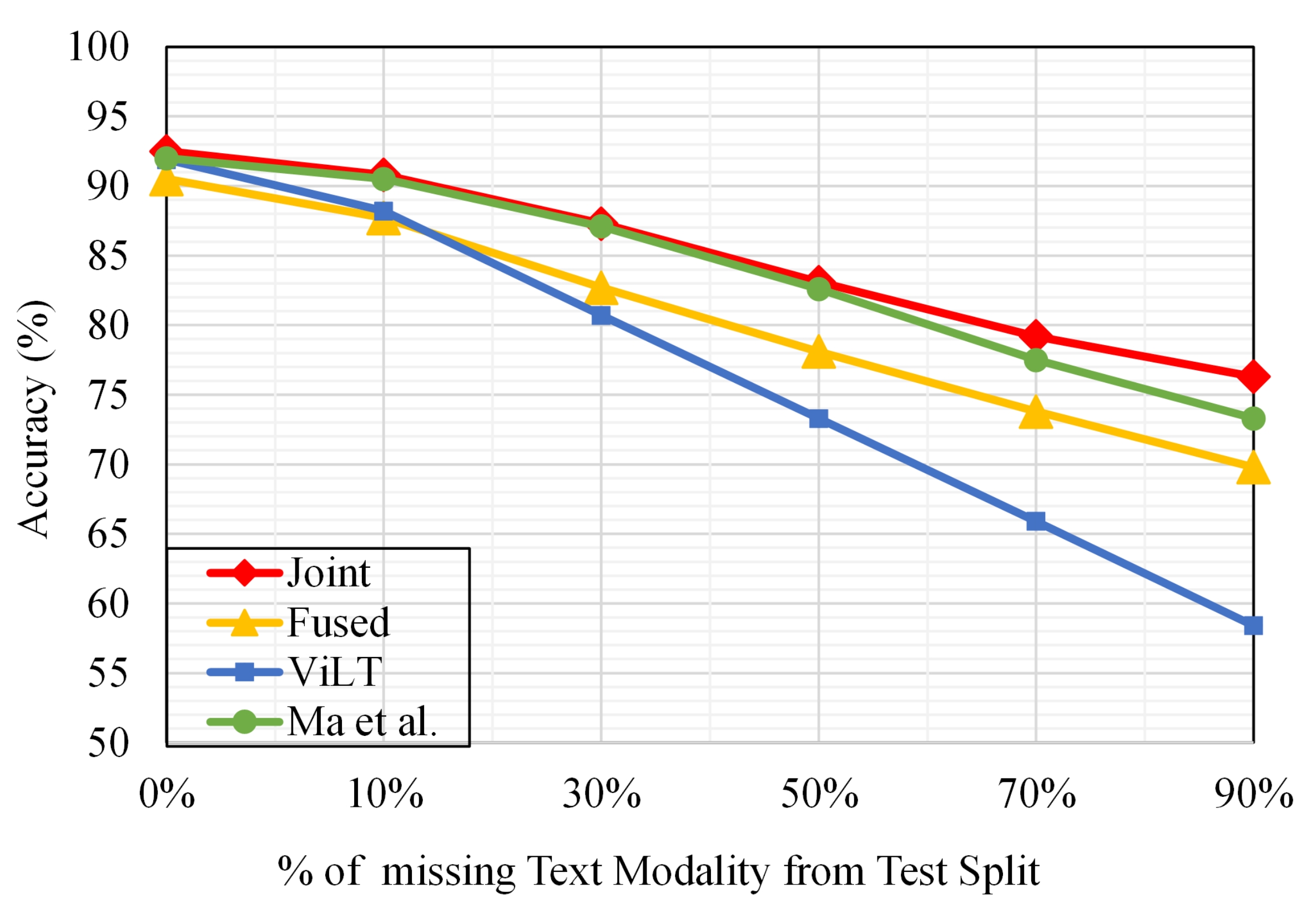}
         \caption{UPMC Food-$101$}
         \label{subfig:upmc}
     \end{subfigure}
     \begin{subfigure}[b]{0.44\textwidth}
         \centering
         \includegraphics[width=\textwidth]{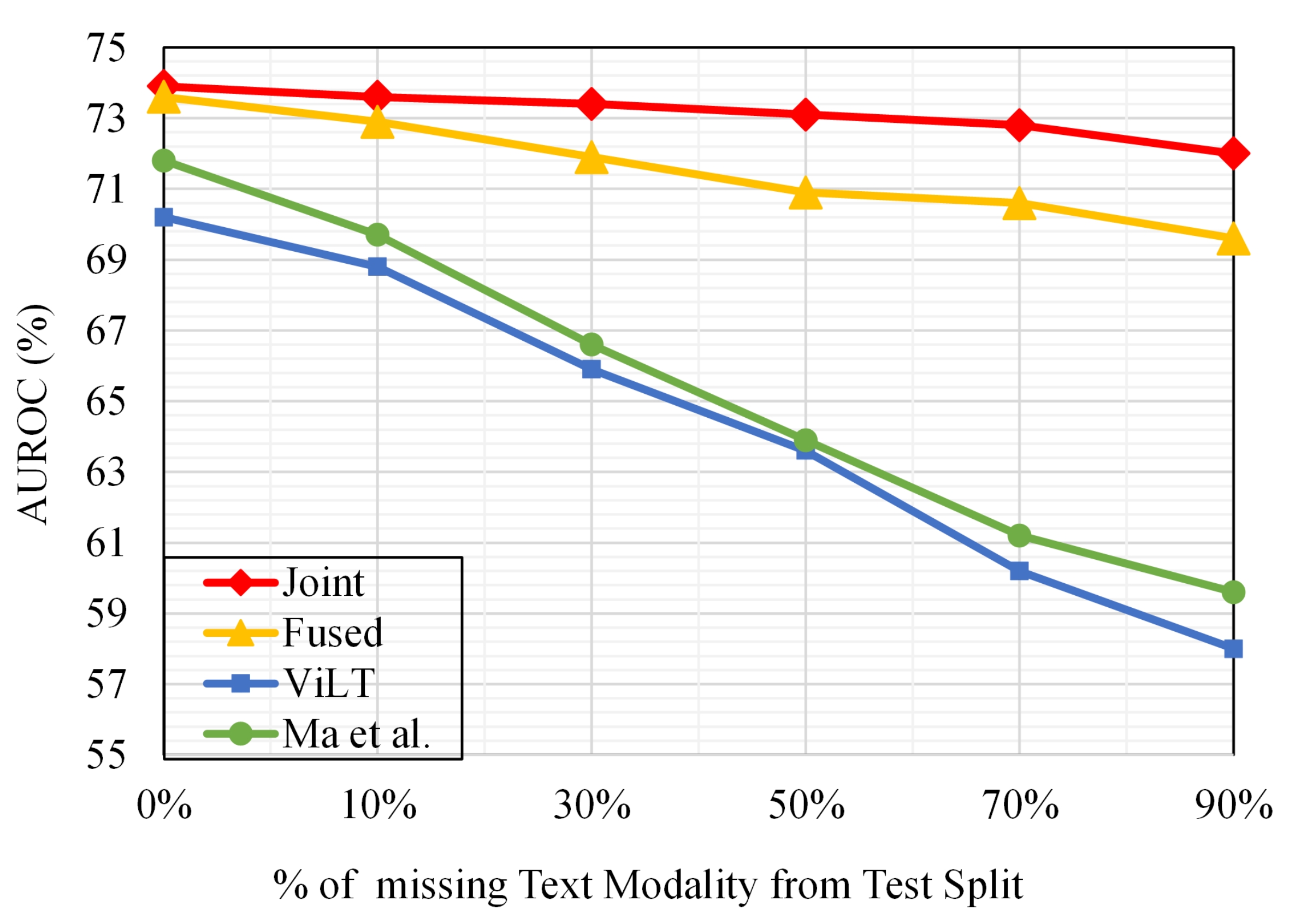}
         \caption{Hateful Memes}
         \label{subfig:hateful}
     \end{subfigure}
        \caption{Performance comparisons of the two variants of \our{} (\textit{fused} and \textit{joint}) with ViLT~\cite{kim2021vilt} and Ma et al. \cite{ma2022multimodal} on various levels of missing textual modality during testing on UPMC Food-101 and Hateful Memes datasets. 
        A smaller drop in performance by our approach in most cases signifies its effectiveness towards training Vision Transformers resilient to missing modalities without dataset-centric fusion strategies.}
        \label{fig:fusion}
\end{figure}

\subsection{Analysis and Discussion}
\label{subsec:analysis}
In this section, we provide further analysis on \our{} including its application to different visual networks, dataset independence, and visualizations for further insights.  

\subsubsection{Is Chameleon Agnostic to the Visual Network?}
\label{subsec:modeagnostic}
The encoding scheme of \our{} makes it generally usable to train any visual network. 
Investigating the extent to which the visual classification network and text encoding method influence performance, we conduct a series of experiments, using ViT and CNN (ResNet-$101$) for the former and 
\textit{joint} and \textit{fused} variants 
for the latter. 
The results of these experiments are summarized in Table~\ref{tab:network_space}. As seen, although Transformer with joint encoding outperforms all other variants of our approach, RestNet-$101$ demonstrates comparable performance. An interesting observation here is that the \textit{fused} approach works slightly better in the case of CNNs whereas \textit{joint} works slightly better in the case of ViTs. Overall, the comparable performance of ViTs and CNNs demonstrates that \our{} is model-agnostic and may be applied to CNNs as well as ViTs. 

\subsubsection{Are Dataset-specific Transformers Necessary?}
\label{subsec:jointvsfused}
Ma et al.~\cite{ma2022multimodal} have observed that multimodal Transformers are not only sensitive to missing modalities but different design choices, such as fusion strategy, should be tailored to the dataset. 
They further conclude that it may not be possible to design a general multimodal Transformer architecture to be used across datasets.
In contrast, with our proposed design of encoding textual modality into visual format and then subsequently training a visual network such as ViT, \our{} becomes completely independent of dataset-centric design choices. 
To evaluate this, we train two separate ViTs using both \textit{fused} and \textit{joint} variants of \our{} on the UPMC Food-$101$ and Hateful Memes datasets. Each model is then evaluated on different levels of missing textual modality at test time. Fig.~\ref{fig:fusion} shows the corresponding results and compares the performances with ViLT \cite{kim2021vilt} and Ma et al. \cite{ma2022multimodal}. 
As Ma et al. have proposed dataset-optimal fusion strategies, their approach demonstrates  robustness to missing modality compared to ViLT. \our{}, on the other hand, in both variants, demonstrates superior resilience against missing modality while training on similar ViTs without any changes across datasets. Our study suggests that \our{} enables the training of dataset-agnostic transformers capable of handling severe missing modalities.

\begin{table}[t]
\caption{Performance analysis of fused and joint representation with different visual networks using UPMC-Food-$101$ and Hateful Memes datasets. 
AUROC and accuracy are reported for Hateful Memes and UPMC-Food-$101$, respectively.}
\centering
\footnotesize
\begin{tabular}{c|c|c||c}
\hline
Dataset (Performance Metric) & Method & ResNet-$101$ & ViT \\
\hline\hline
\multirow{2}{*}{UPMC Food-$101$ (Accuracy)} & Fused & 91.2 & 90.5 \\ 
\cline{2-4}
                                            & Joint & 89.5 & \textbf{92.5} \\
\hline\hline
\multirow{2}{*}{Hateful Memes (AUROC)}      & Fused & 71.2 & 73.6 \\ 
\cline{2-4}
                                            & Joint & 70.9 & \textbf{73.9} \\
\hline
\end{tabular}
\label{tab:network_space}
\end{table}

\begin{figure*}[t]
\centering
\includegraphics[scale=0.3]{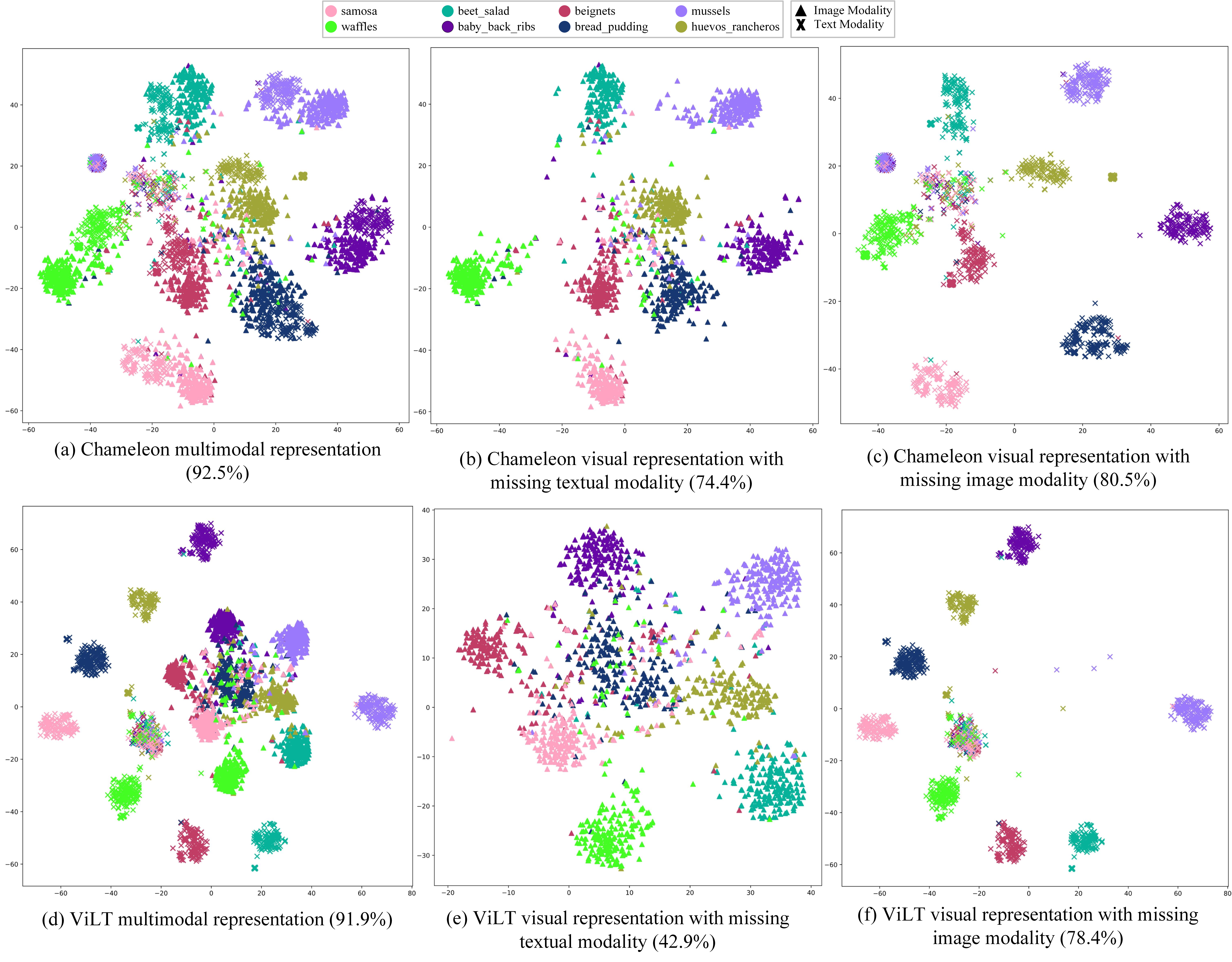}
\vspace{-2mm}
\caption{t-SNE visualizations of the embedding space of \our{} (a - c) and ViLT (d - f) along with accuracy on test set of UPMC Food-$101$.
Compared to ViLT, \our{} not only enhances the classification boundaries when complete modalities are available at test time but also retains these boundaries when the textual or visual modality is completely missing during test time.
Note that classes are selected randomly from the test set.
}
\label{fig:ntsne}
\end{figure*}

\subsubsection{t-SNE Visualizations.}
We plot results of t-SNE projections to take a peek into the embedding space of \our{} trained on UPMC Food-$101$ dataset and provide comparisons with ViLT~\cite{kim2021vilt} in Fig. \ref{fig:ntsne}.
Comparisons are provided under the following three settings: Fig.~\ref{fig:ntsne}a \& d) complete modalities during training and testing, b \& e) complete modalities during training but $100$\% missing textual modality during testing, and c \& f) complete modalities during training but $100$\% missing visual modality during testing. 
In the case of multimodal training/testing, compared to ViLT, \our{} is able to project modalities belonging to the same classes closer to each other while demonstrating inter-class separability (Fig.~\ref{fig:ntsne}a \& d). 
This highlights the success of \our{} in training a robust multimodal classifier.
Finally, in the case of multimodal training but 100\% missing modality during testing (Fig.~\ref{fig:ntsne}b, c, e \& f), although some distortions are noticeable, compared to ViLT, \our{} successfully retains the overall separability of the classes under such severe missing modality case.
This demonstrates the resilience of \our{} towards missing modalities.

\subsection{Limitations}

While advantageous in most missing modality cases, utilizing text as an image has its limitations as well. For example, in the case of longer text descriptions, the network struggles to learn joint representation of multiple modalities as well as achieving robustness to missing modalities. This phenomenon can be observed in Tables~\ref{tab:percentage_missing} \& \ref{tab:traing_missing} where CLIPPO, utilizing rendered text as an image, performs notably lower than its counterpart networks. In the case of \our{}, this problem is partially handled as we convert the textual embedding into encoded visual representations before training a network. This way, longer texts are summarized into compact encoded visual representations thus yielding superior performance.



\section{Conclusion}
\label{sec:conclusion}
In this work, we presented a robust textual-visual multimodal learning method that encodes word embeddings of the textual modality into visual
format to learn modality-independent representations of the input modalities.
The common input format facilitates joint representation learning by sharing weights across multiple modalities resulting in a multimodal network robust to missing modalities.
Extensive experiments are performed on the popular and challenging datasets UPMC Food-$101$, Hateful Memes, MM-IMDb and Ferramenta.  
The proposed method is thoroughly evaluated on complete modalities as well as missing modalities during training and testing.  
The experimental results indicate that the proposed method obtains superior performance when a complete set of modalities is available. 
In the case of missing modalities, the performance deterioration is noticeably smaller than that of the existing multimodal learning methods, indicating significant robustness of \our{}.


%
%
\bibliographystyle{splncs04}
\bibliography{main}
\end{document}